# Tackling Data Scarcity with Transfer Learning: A Case Study of Thickness Characterization from Optical Spectra of Perovskite Thin Films

Siyu Isaac Parker Tian[1,2], Zekun Ren[1,2,†], Selvaraj Venkataraj[2], Yuanhang Cheng[2,‡], Daniil Bash[3], Felipe Oviedo[4,§], J. Senthilnath[5], Vijila Chellappan[3], Yee-Fun Lim[3,6], Armin G. Aberle[2], Benjamin P MacLeod[7], Fraser G. L. Parlane[7], Curtis P. Berlinguette[7], Qianxiao Li[8], Tonio Buonassisi[1,4,*], Zhe Liu[1,4,*,l]

[1] Low Energy Electronic Systems (LEES), Singapore-MIT Alliance for Research and Technology (SMART), 1 Create Way, Singapore 138602, Singapore

[2] Solar Energy Research Institute of Singapore (SERIS), National University of Singapore, 7 Engineering Drive, Singapore 117574, Singapore

[3] Institute of Materials Research and Engineering (IMRE), Agency for Science, Technology and Research (A*STAR), 2 Fusionopolis Way, Singapore 138634, Singapore

[4] Department of Mechanical Engineering, Massachusetts Institute of Technology (MIT), 77 Massachusetts Ave., Cambridge, MA 02139, USA

[5] Institute for Infocomm Research (I2R), Agency for Science, Technology and Research (A*STAR), 1 Fusionopolis Way, Singapore 138632, Singapore

[6] Institute of Sustainability for Chemicals, Energy and Environment, Agency for Science, Technology and Research (A*STAR), 1 Pesek Rd, Singapore 627833, Singapore

[7] Department of Chemistry, The University of British Columbia (UBC), 2036 Main Mall, Vancouver, BC V6T 1Z1, Canada

[8] Department of Mathematics, National University of Singapore (NUS), 21 Lower Kent Ridge Rd, Singapore 119077, Singapore

[†] Now at: Xinterra, Singapore, 77 Robinson Road, Singapore 068896, Singapore

[‡] Now at: Department of Materials Science and Engineering, City University of Hong Kong, Kowloon, Hong Kong, 999077, P.R. China

[§] Now at: Microsoft AI for Good, Redmond, WA 98052, USA

[l] Now at: School of Materials Science and Engineering, Northwestern Polytechnical University, Xi'an, Shaanxi, 710072, P.R. China

[*] Corresponding authors: Z.L. (zhe.liu@nwpu.edu.cn) and T.B. (buonassi@mit.edu)

**Abstract**

Transfer learning increasingly becomes an important tool in handling data scarcity often encountered in machine learning. In the application of high-throughput thickness as a downstream process of the high-throughput optimization of optoelectronic thin films with autonomous workflows, data scarcity occurs especially for new materials. To achieve high-throughput thickness characterization, we propose a machine learning model called *thicknessML* that predicts thickness from UV-Vis spectrophotometry input and an overarching transfer learning workflow. We demonstrate the transfer learning workflow from generic source domain of generic band-gapped materials to specific target domain of perovskite materials, where the target domain data only come from limited number (18) of refractive indices from literature. The target domain can be easily extended to other material classes with a few literature data. Defining thickness prediction accuracy to be within-10% deviation, *thicknessML* achieves 92.2 ± 3.6% accuracy with transfer learning compared to 81.8 ± 11.7% without (lower mean and larger standard deviation). Experimental validation on six deposited perovskite films also corroborates the efficacy of the proposed workflow by yielding a 10.5% mean absolute percentage error (MAPE).

# I. Introduction

The recent advances in robotic automation in research laboratories have enabled autonomous high-throughput experimentation (HTE) workflows for synthesis, screening, and optimization of new materials [1]–[10]. These HTE workflows can generate new materials in thin-film form at a record rate (*e.g.*, a few minutes per sample) [11]–[13], and therefore materials characterization and data analysis at downstream must match the elevated throughput. To accelerate data analysis and knowledge extraction in HTE, ML algorithms are used together with rapid characterization techniques [14]–[20]. For materials in thin-film form, film thickness is among the most essential and yet challenging parameters to measure in high-throughput non-destructive manner [21]–[26].

The state-of-the-art characterization method is optical spectroscopy. Despite its rapid measurement, a manual fitting of optical models for the parametric description of the material's wavelength-resolved refractive indices must ensue to obtain thickness. However, this manual fitting for a new material is consequently a bit slow ranging from a few tens of minutes to hours per sample, and it usually requires much experience on top of trial and error. The refractive indices of different material classes fall into different distributions (domains), reflected by the different numbers and types of optical models typically used for different material classes. For each specific domain (material class), especially of newly developed materials such as lead-halide perovskites, readily available refractive indices data are few, and we face the bottleneck of data scarcity.

The use of transfer learning has been gradually rising nowadays in materials science to counter data scarcity prevalently present in many applications. Notable examples lie heavily in materials property prediction, where the learning transfers across properties [27]–[32], across modes of observation, *e.g.*, from calculated properties to experimental ones [27], [30], and across different materials systems [31], *e.g.*, from inorganic materials to organic polymers [30], or from alloys to high-entropy alloys [32]. Following the same rationale, thin film thickness characterization also presents itself as a suitable field for the application of transfer learning to overcome its bottleneck of data scarcity. Thickness prediction in the target domain of perovskite films is just a case study of this transfer learning workflow. The target domain can be easily extended to other material classes. The requirement is simply to find a few experimentally fitted refractive indices from literature of that specific material class.

In this work, as a case study to demonstrate high-throughput thickness characterization with ML in certain domain (with generality to extend to other domains), we propose the following transfer learning workflow (Figure 1) to automatically extract film thickness from optical spectra with limited data. We use an example of lead-halide perovskites, which is a family of $ABX_3$ semiconductors with excellent optoelectronic properties, *e.g.*, for photovoltaics, light-emitting diodes, photodetectors. The ML workflow centers on a model named *thicknessML* that takes optical reflection ($R$) and transmission ($T$) spectra as input and outputs thickness ($d$) and wavelength-resolved refractive indices (Figure 1a). We denote the real and imaginary parts of refractive indices as $n$ and $k$ respectively.

Transfer learning happens from the source domain to the target domain (Figure 1b). The source domain contains generic semiconductor refractive indices — we achieve this by parametrically simulating refractive indices from a single optical model (Tauc-Lorentz) commonly used for optical materials with an absorption bandgap. The source domain is considered "big data" due to simulation. The target domain contains perovskite semiconductor refractive indices that are experimentally fitted and scarce and thus considered "small data". In practice, we find 18 perovskite refractive indices from literature to form our target domain. A two-stage training occurs to transfer from the source to the target domain — I) pre-train *thicknessML* on the source domain, II) retrain *thicknessML* on the target domain. As a result, *thicknessML* attains a more accurate and precise mean absolute percentage error (MAPE) of 4.6 ± 0.5% for thickness prediction with limited perovskite data, compared to 7.4 ± 4.2% for the direct training. This is quantified on target-domain data derived from a training-test split of 13-5 literature refractive indices. An experimental validation of *thicknessML* on six deposited methylammonium lead iodide (MAPbI$_3$) perovskite films yields 10.5% MAPE from retraining on target-domain data derived from eight dissimilar literature refractive indices whose perovskite compositions contain no methylammonium.

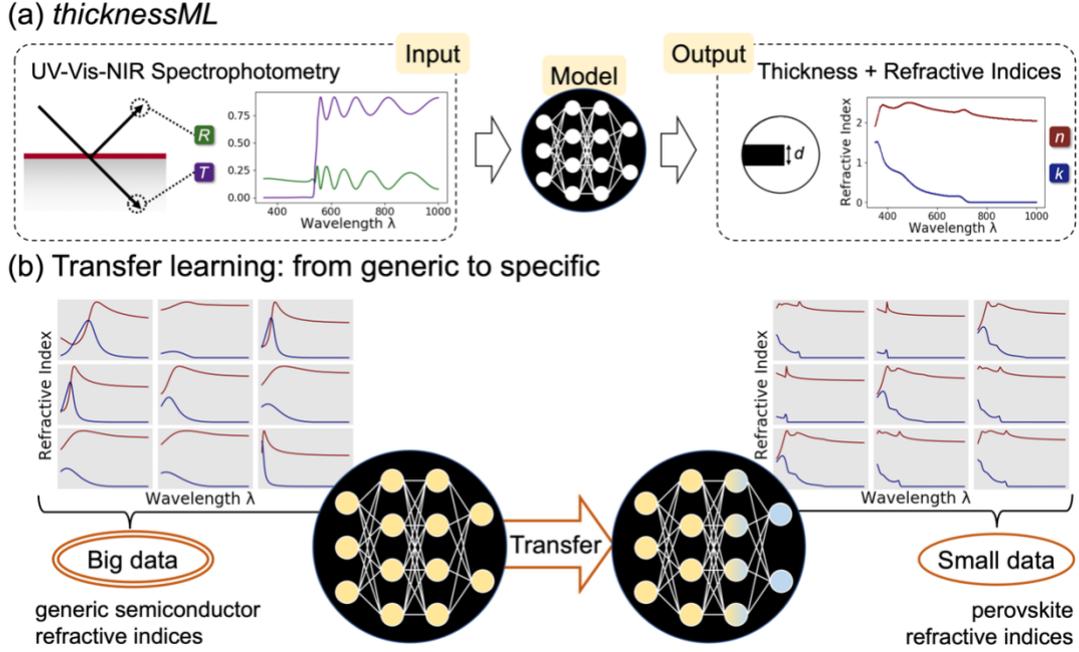

Figure 1. Transfer learning workflow. **a** Inputs and outputs of the thickness extraction model named *thicknessML*. $R$ and $T$ are reflectance and transmittance resulting from UV-Vis-NIR spectrophotometry. $d$ is thickness. $n$ and $k$ are real and imaginary parts of the refractive index. **b** Transfer learning stages — from generic (source domain) to specific (target domain). The source domain contains generic semiconductor refractive indices (simulated and thus of big data). The target domain contains specific (perovskite) refractive indices (experimentally fitted and thus of small data). *thicknessML* is first pre-trained in the source domain, and then transferred to (retrained in) the target domain.

## II. Results and Discussion

**Preparation of Source and Target Datasets**

For ML datasets, the inputs are wavelength-resolved optical spectra of reflection and transmission. Denoting wavelength as $\lambda$, the optical spectra are respectively $R(\lambda)$ and $T(\lambda)$ for reflection and transmission. The outputs (or labels in supervised ML terminology) are thickness $d$, and wavelength-resolved refractive indices $n(\lambda)$ and $k(\lambda)$. The refractive index is complex, and $n$ and $k$ denote the real and imaginary parts respectively, namely $\tilde{n}(\lambda) = n(\lambda) + ik(\lambda)$. In physics, the order is reverse, where thickness $d$ (an extensive property of a material) and the refractive index (an intensive property of a material) are inputs, and the optical spectra are outputs (an optical response given a material film). Consequently, the ML model in essence is learning the inverse of the physical optical response, and the physical response is universal for all materials (across material classes). The source / target domains (material classes) only appear due to different underlying distributions of an output (label), namely the refractive index. The different refractive index distributions are a manifestation of the different underlying governing parametric optical models in different material classes.

To facilitate transfer learning, we built the source dataset to be generic; practically, we simulated refractive indices with Tauc-Lorentz (TL) optical model, universal for materials with a band gap. The simulation of refractive indices is analogous to the simulation of materials (possessing the simulated refractive index). Paired with different thicknesses, a set of simulated refractive index spectra (a simulated material) can yield the respective optical spectra through the optical response. This is analogous to taking the optical response of a batch of thin films (different thicknesses) made of the same material (the same simulated refractive index spectra). The optical response was simulated by the physical transfer-matrix method (TMM). Without loss of generality, we adopted 0° incident angle, and 1 mm glass substrate in the TMM simulation.

In the source dataset, we simulated 1,116 $n(\lambda)$, $k(\lambda)$ spectra by sampling a grid of parameter values (for $A$, $C$, $E_0$, $E_g$, with a fixed $\varepsilon_\infty=1$) in a Python implementation of a single TL optical model with $\lambda$ ranging from 350 to 1000 nm. The $\lambda$ range was chosen to be a common subset of frequent ranges in UV-Vis measurements and reported literature. The 1,116 $n$, $k$ spectra of the source dataset were divided into 702, 302, and 112 for the training, validation and test set respectively as shown in Table 1. Then we randomly chose 10 thicknesses per pair of $n$, $k$ spectra (per simulated material) in the training and validation set, and 50 thicknesses per pair in the test set to obtain corresponding $R$, $T$ spectra in the training, validation and test set. The larger number of $d$ per pair of $n$, $k$ spectra in the test set gives a more stringent thus reliable evaluation of how well *thicknessML* behaves. The range of $d$ is 10 – 2010 nm. Three different splits were performed to the training, validation, and test set for three ensemble runs, and the randomly selected thicknesses for the same $n$, $k$ spectra also differed in the three splits.

Table 1. Training, validation, and test set in the source dataset

|  | Training set | Validation set | Test set |
| --- | --- | --- | --- |
| Number of $n$, $k$ spectra | 702 | 302 | 112 |
| Number of $d$ per $n$, $k$ spectra | 10 | 10 | 50 |
| Resultng number of $R$, $T$ spectra | 7020 | 3020 | 5600 |

In the target dataset, we procured 18 perovskite $n(\lambda)$, $k(\lambda)$ spectra from literature [33]–[37]. Instead of one training-test split (validation set not used in the target dataset due to data scarcity), we evaluated the performance of transfer learning on increasing number of training $n$, $k$ spectra from 0 to 17, with the corresponding rest in the test set. We maintained the number of $d$ per $n$, $k$ spectra in the training (10) and test set (50) to obtain resulting $R$, $T$ spectra. To compare with training from scratch (direct training without transfer), we evaluated one case, a 13–5 training-test split (for number of $n$, $k$ spectra), and assigned 500 $d$ per $n$, $k$ spectra for training while maintaining the 50 $d$ per $n$, $k$ spectra for test. The extremely large number of assigned $d$ per $n$, $k$ spectra was to complement the small number of $n$, $k$ spectra available, ensuring enough training data in learning from scratch. Five training-test splits were performed for the ensemble runs.

**Model: *thicknessML***

The whole framework of *thicknessML* is shown in Figure 2. For *thicknessML*, we propose a convolutional neural network (CNN) architecture [38]–[40]. CNNs are originally developed for image processing and can capture local spatial information (neighboring pixels) as well as correlation among channels, such as RGB as channels. We used a CNN here to capture $R$, and $T$ segments at contiguous wavelengths, and thus capture some spatial features like hills and valleys of $R(\lambda)$ and $T(\lambda)$, which are closely related to thickness $d$. We also concatenated $R$, and $T$ channel-wise to capture the correlation between $R$, and $T$ in accordance with the Kramers–Kronig relations in physics [41].

Aside from a straightforward *RT*-to-*d* architecture, we also explored a multitask learning (MTL) architecture, where $n(\lambda)$ and $k(\lambda)$ are also outputs together with $d$. This is inspired from the fact that the physics in determining $d$ from $R$ and $T$ is highly related to the concurrent determination of $n$ and $k$. Therefore, MTL, concurrent learning of multiple tasks, is a suitable architecture to capture this relation. In MTL, if the tasks are related, the multiple tasks can help the learning to be more accurate, and less likely to overfit to a specific task (in other word, more generalized learning) [42], [43]. As a result, we concurrently learnt to predict $d$ as our main task, and $n$, $k$ as auxiliary tasks in our MTL implementation. The straightforward *RT*-to-*d* architecture is then single task learning (STL) without the auxiliary tasks, directly satisfying our goal—thickness extraction.

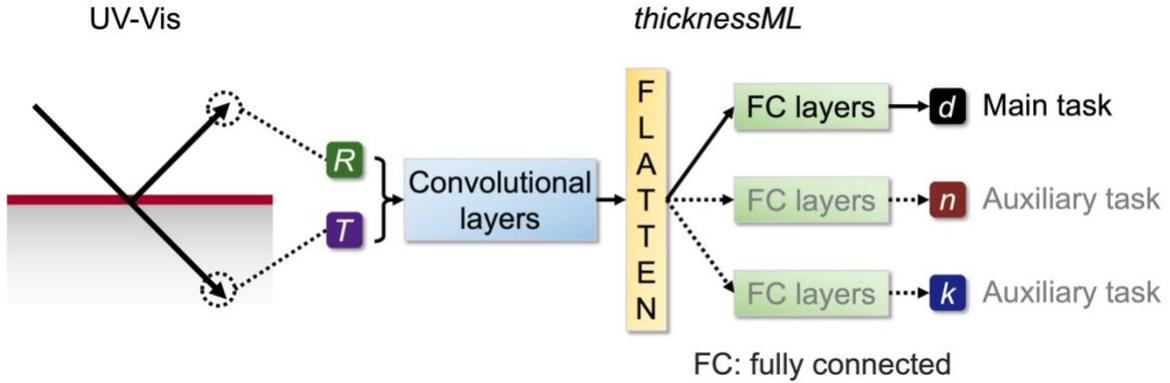

Figure 2. *thicknessML* framework: *thicknessML* receives the $R(\lambda)$, and $T(\lambda)$ spectra and outputs $d$ (and $n(\lambda)$, $k(\lambda)$) for **S**ingle **T**ask **L**earning (**M**ulti**T**ask **L**earning). Input $R$, and $T$ spectra first go through four convolutional and max pooling layers for feature extractions, and then get flattened to be passed to three fully connected (FC) and dropout layers, where mapping from extracted features to task targets are drawn. The three dedicated FC-layer blocks for $d$, $n(\lambda)$, and $k(\lambda)$ corresponds to MTL implementation. STL implementation has the same architecture without the two FC-layer branches for $n(\lambda)$, and $k(\lambda)$. (The adopted incident angle in the UV-Vis is 0°. The inclined beams are drawn to achieve better visual clarity.) The detailed hyperparameters is recorded in S1 section of Supplementary Information.

The term "*thicknessML*" has both specific and general definitions, depending on context. Specifically, it refers to the ML model that outputs thickness (and refractive index) from optical $R$ and $T$ data. Generally, it encompasses the whole of the thickness extraction method, including UV-Vis operation. In this study, we also use the terms "thickness extraction" and "thickness prediction" interchangeably.

**Stage I: Pre-training on the generic source dataset**

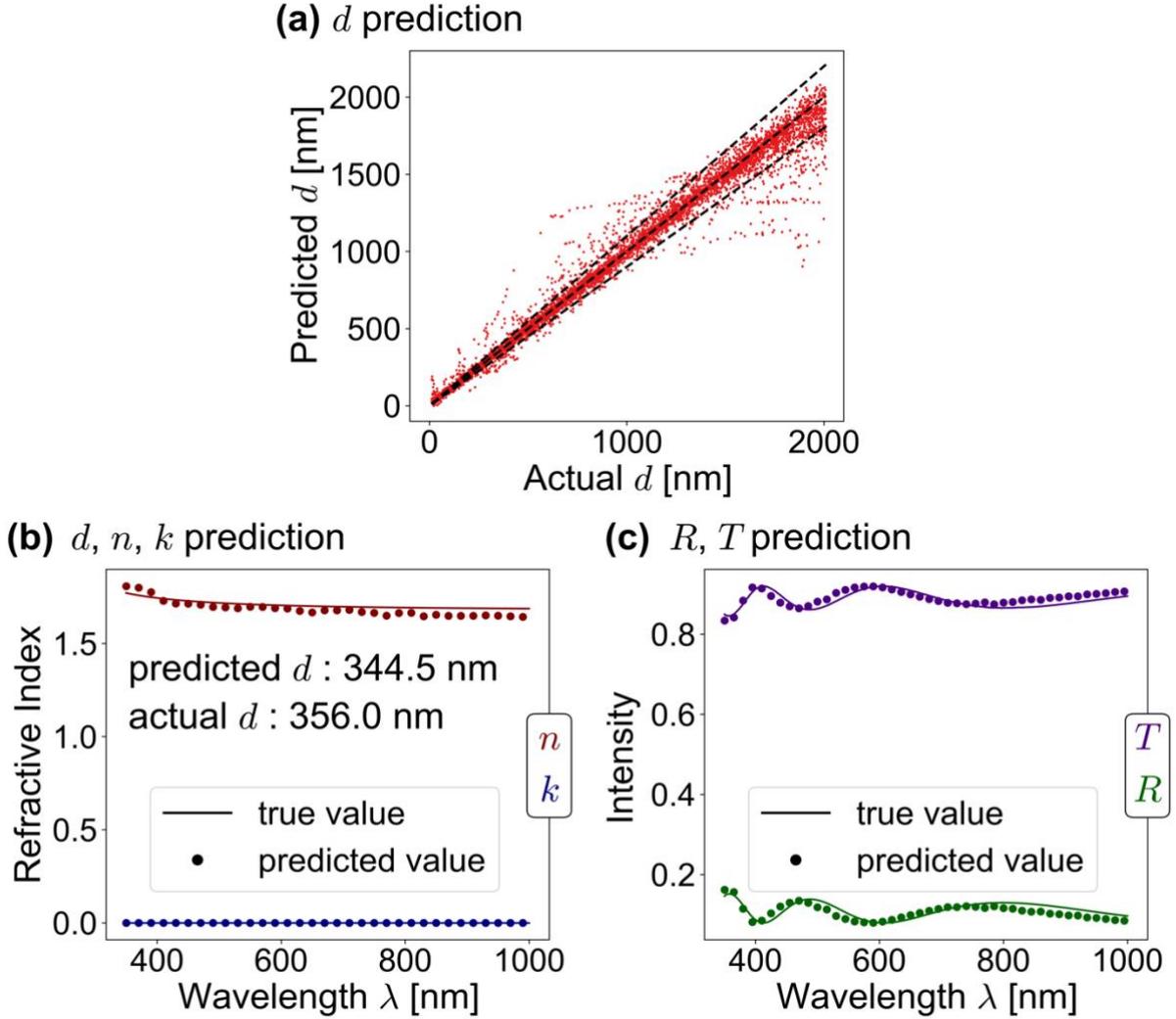

Figure 3. Performance of *thicknessML* (MTL) on the test set (the best run out of three ensemble runs) **a** Predicted *d* vs. actual *d*, where the diagnal line indicates perfect prediction, and the two side lines ±10% deviation. **b** *d*, *n*, *k* prediction of an arbitrary sample, where dots denote predictions, and lines actual values. (Dots of 5 nm $\lambda$ increment are plotted for better visual clarity when predictions are with 1 nm increment as the actual *n*, *k* spectra.) **c** *R*, *T* reconstruction from predicted *d*, *n*, *k* using TMM on top of actual input *R*, *T* spectra, where dots denote predictions, and lines actual values.

Table 2. Performance comparison of *thicknessML* (averaged of three ensemble runs)

|  | *d* (<10% deviation) | *n* (<10% deviation) | *k* (<10% deviation) |
|---|---|---|---|
| STL | 89.2% |  |  |
| MTL | 83.3% | 94.2% | 26.8% |

To evaluate the performance of our pre-trained *thicknessML* performance, we show the results on the generic source dataset (the best run out of the three ensemble runs) in Figure 3. We also illustrate the criteria used for performance evaluation. Three different train-validation-test splits with respect to *n*, *k* spectra are carried out for both STL and MTL. The results from the best performing MTL run (in terms of thickness) were selected and shown in Figure 3. Comparison of performances of STL and MTL is documented in Table 2, where the performances were averaged across three ensemble runs.

The criterion adopted stands out most perceivably in *d* prediction in Figure 3a, where we quantify the percentage of predictions falling within 10% deviations from the actual values, *i.e.*, the proportion of the points falling within the two side diagonal lines denoting 10% deviation from perfect predictions. Similar criteria pervade through the *n*, *k* prediction evaluation in MTL: since *n*, and *k* are wavelength-resolved, the within-10%-deviation criterion is quantified across wavelengths on average. For instance, in Figure 3b *n* and *k* predictions denoted by the dots should on average across wavelengths fall within 10% from the actual *n*, *k* spectra denoted by lines.

Table 2 summarizes the performances. For *d* prediction, STL achieves 89.2% accuracy (5.0% MAPE), and MTL 83.3% (8.0% MAPE). STL slightly betters MTL, seemingly contradicting our perception that MTL promotes more generalized learning and better accuracy. This result brings forth a trade-off between a generalized learning and a task-specific learning: a generalized learning may improve the performance through learning more generalized features of the input, but also may negatively impact the performance by scattering learning capacity across various tasks (thus losing focus on the intended main task). Accuracies reach 94.2% and 26.8% for *n* and *k* prediction respectively in MTL. We notice the relatively poor performance of *k* prediction, and we attribute it to several reasons:

- Many *k* values on larger wavelengths are near or at zero, *e.g.*, on the magnitude of $10^{-2}$. This renders the percentage-based within-10%-deviation criterion unduly stringent. A different choice of absolute-error-based criterion may reflect the *k* prediction performance more appropriately.
- The many near-zero and at-zero *k* values bias the output data distribution heavily and unfavorably.
- The prediction of wavelength-resolved values is naturally harder than the prediction of a scalar value.

Nevertheless, we acknowledge the *k* prediction limitation of *thicknessML*-MTL, and caution potential users to place more confidence in the *d* prediction than the *n*, *k* prediction, as the inclusion of auxiliary tasks (*n*, *k* prediction) is to help the main task, *d* prediction.

**Stage II: Transfer learning to the perovskite target dataset**

Transfer learning takes the pre-trained model (a warm start from the pre-trained weights instead of random initializations) and let the partial or full weights continue to train on a new dataset (retraining). Through the pre-trained weights, transfer learning [44] allows knowledge learnt in the pre-trained task to be transferred to a related new task with much less data and training. In our case, the pre-trained weights carry the knowledge of an inverse mapping of TMM, from *R*, and *T* to *d*, *n*, and *k*, and the retraining further adapts the mapping to a dataset whose underlying oscillator models are of perovskite materials.

We propose and run two types of transfer learning: (1) full-weight retraining, to continue updating the weights for both convolutional and fully connected layers. (2) partial-weight retraining, to freeze the weights for convolutional layers, responsible for feature extraction, while updating the fully connected layers. To provide a baseline, we also implemented a case of direct training / learning from scratch (from random initialized weights as in pre-training).

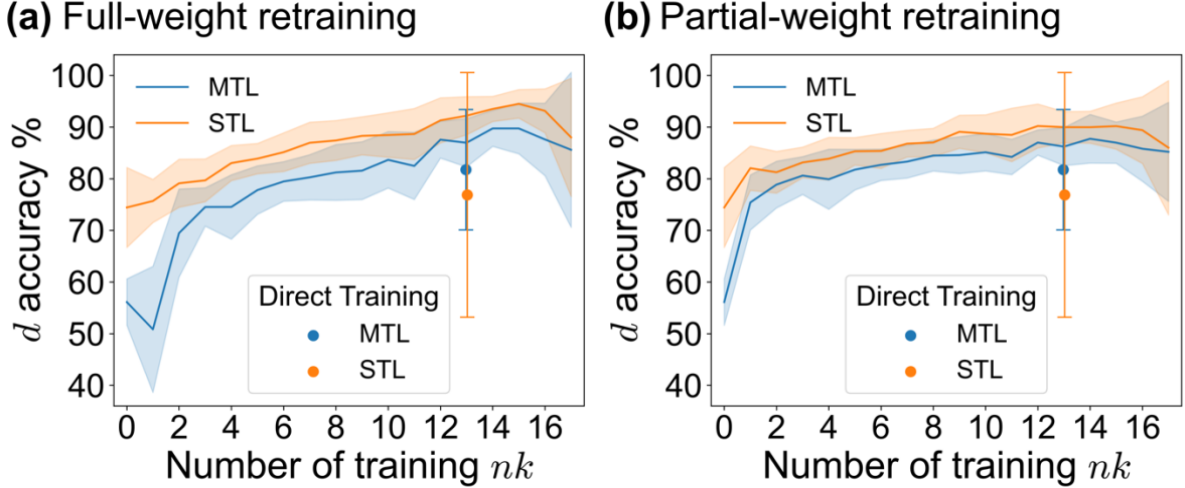

Figure 4. $d$ accuracy (evaluated on the test set) vs. number of retraining $n$, $k$ spectra (out of a total of 18) **a** Full-weight retraining, where every weight is updated during retraining. **b** Partial-weight retraining, where weights in the convolutional layers are frozen, and only the weights in fully connected layers are updated. Bands denote transfer learning using the corresponding number of retraining $n$, $k$ spectra with 10 $d$ per $n$, $k$ spectra; solid lines and spreads denote the mean, and the standard deviation of runs from 5 different train-test splits and an ensemble of three pre-trained *thicknessML* models; dots with error bars denote a case of direct training on 13 $n$, $k$ spectra with 500 $d$ per n, k spectra, while the dot and the error bar denote the mean, and the standard deviation of runs from 5 different train-test splits.

Figure 4 shows the transfer learning results compared with direct training on the perovskite target dataset and offers several observations. Firstly, transfer learning achieves better accuracy (higher mean) and precision (smaller spread) than direct training in either case. Although able to exceed the performance of transfer learning in certain individual runs, direct training seems to be largely affected by specific train-test splits. Transfer learning also requires much less data compared to direct training (10 $d$ per $n$, $k$ spectra vs. 500 $d$ per $n$, $k$ spectra). These results demonstrate the superiority of transfer learning and the efficacy of the transfer learning workflow in the thickness prediction of thin films. Secondly, partial-weight retraining behaves better when the number of retraining $n$, $k$ spectra is small ($\leq 11$). Full-weight retraining performs better when the number of retraining $n$, $k$ spectra is large ($>11$). Full-weight retraining experiences a decrease in accuracy first, then followed by an eventual increase. Compared to partial-weight retraining, full-weight retraining has more weights to update, and thus is more flexible. Flexibility has both pros and cons: when the number of retraining $n$, $k$ data is small, flexibility more easily steers *thicknessML* away from optimal weights (an initial drop in accuracy); when the number of retraining $n$, $k$ spectra becomes large enough, flexibility offers a higher learning capacity, and thus a better accuracy. Lastly, STL slightly betters MTL in both kinds of transfer learning, while MTL gives more accurate (higher mean) and precise (smaller spread) direct training results. Again, we attribute this comparison between STL and MTL to the trade-off between learning more generalized features and less focused learning (on the main task). Overall, we recommend the STL implementation in the transfer learning workflow paired with either partial-weight retraining (when the number of retraining $n$, $k$ is smaller) or full-weight retraining (when the number of retraining $n$, $k$ is larger). We follow this recommendation in our ensuing experimental validation.

To quantify the improvement of transfer learning compared to direct training, we report the transfer learning and direct training results in Table 3.

Table 3. Transfer learning *vs.* direct training results. Only the better-performing results are reported between MTL and STL implementations. The results are recorded in a format of mean $\pm$ standard deviation. The best performing results in each category, *i.e.*, transfer learning and direct training, are in bold.

|  | Transfer Learning | | Direct Training |
| --- | --- | --- | --- |
|  | Full-weight retraining | Partial-weight retraining |  |
| *d* accuracy* | **92.2 $\pm$ 3.6%** (STL) | 90.0 $\pm$ 2.9% (STL) | **81.8 $\pm$ 11.7%** (MTL) |
| *d* MAPE | **4.6 $\pm$ 0.5%** (STL) | 4.9 $\pm$ 0.6% (STL) | **7.4 $\pm$ 4.2%** (MTL) |

*This is the percentage of accurately predicted *d*, where accuracy is defined as within 10% deviation.

**Experimental validation with six experimental perovskite thin films**

To validate this transfer learning workflow of *thicknessML*, we validated its performance on experimental perovskite films. We deposited six methylammonium lead iodide (MAPbI$_3$) films with assorted precursor solution concentrations and spin coating speeds recorded in Table 3. We performed UV-Vis and profilometry measurements and compared them with the *thicknessML* predictions in Figure 5. Pre-trained *thicknessML* was retrained on the perovskite target dataset with eight retraining *n*, *k* spectra using a partial-weight retraining. The choice of eight perovskites was deliberate to be more distinct from MAPbI$_3$ by not containing methylammonium (MA). The resulting predictions have a 10.5% MAPE.

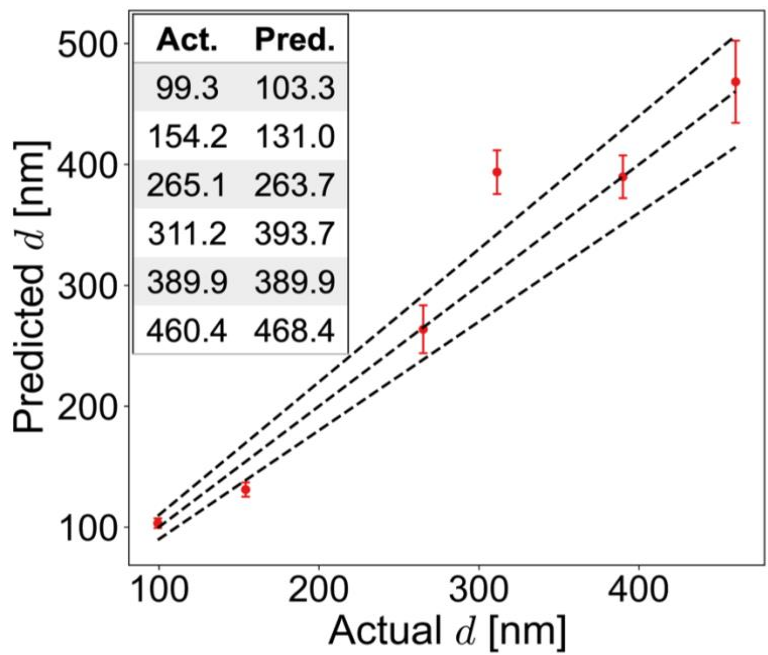

Figure 5. *thicknessML*-predicted thickness vs. profilometry-measured thickness of six perovskite films The inset lists the actual (Act.) measured and predicted (Pred.) thicknesses of the films (mean values for predicted thickness) in unit of nanometer. The error bar on the predicted thickness denotes the standard deviation of various runs of the ensemble *thicknessML*, while the dot the mean. The diagonal line plots the perfect prediction, and the two side lines ±10% deviation.

To evaluate *thicknessML* as a high-throughput characterization framework, we also record its throughput. The bulk of time per sample is spent on UV-Vis, and depends on specific tools, where the prediction of one film only adds about 0.4 ms. In this study, UV-Vis is measured by a stand-alone tool with an integrating sphere and takes about 2 minutes per sample to measure $R(\lambda)$, and $T(\lambda)$ of 0° incident angle. The one-off retraining (transfer learning) of *thicknessML* takes a few minutes, and the one-off pre-training takes about 3.3 hours on a desktop equipped with Intel(R) Core(TM) i7-4790 CPU and NIVIDIA GeForce GTX 1650 GPU.

## III. Conclusions

We used transfer learning to tackle data scarcity in the application of high-throughput thin film thickness characterization. The data scarcity rises from the need to traverse different domains (classes) of materials possessing different underlying optical models while relevant data are scarce for certain domains, especially the domains of newly developed materials such as perovskites. Paired with high-throughput UV-Vis measurement (input), we proposed a transfer learning workflow that first pre-trains on a generic source domain and then transfers to a specific target domain to predict thickness (output). We demonstrated this workflow on the target domain of perovskite materials without loss of generality to extend the target domain to other material classes. On the generic source dataset that is simulated from the generic Tauc-Lorentz optical model, 89.2% of the predicted $d$ from pre-trained *thicknessML* fall within 10% deviation (89.2% $d$ accuracy). After transferring to the specific target dataset that is from 18 literature perovskite refractive indices, retrained *thicknessML* reaches 92.2 ± 3.6% $d$ accuracy compared to 81.8 ± 11.7% $d$ accuracy of direct training. We also validate our transfer learning workflow experimentally on six deposited $MAPbI_3$ films, reaching an MAPE of 10.5%.

Overall, we demonstrated our proposed transfer learning workflow to go from generic to specific, while only the pre-training takes once-off longer training and more data and the transfer learning only took minimal retraining (within minutes) and minimal target data (a few literature refractive indices of materials in the target domain). This workflow can be easily duplicated and extended to other target domains (material classes). We believe that this study opens a new direction of high-throughput thickness characterization and serves as an inspiration for future research encountering data scarcity.

## IV. Methods

**Refractive Index Simulation from single Tauc-Lorentz Oscillator**

In UV-Vis, estimating thickness relies on fitting underlying oscillator models, which describes the interaction between the impingent electromagnetic wave and electrons within the thin film. Specifically, an oscillator model parameterizes the complex wavelength-resolved refractive index $\tilde{n}(\lambda) = n(\lambda) + ik(\lambda)$ via a middleman, the dielectric function, which, with film thickness $d$, determines the optical responses $R(\lambda)$ and $T(\lambda)$. Reflecting the variety of materials and their electron densities of states, there are many types of oscillator models, including Tauc-Lorentz, Cauchy, and Drude, among others [45]–[47]. The TL oscillator, widely used for modeling metal oxides, is a default go-to for modeling materials with band gaps, serving as an indispensable building block for semiconductor optical models. Thus, we choose a single TL oscillator to simulate our generic source dataset.

The Python implementation of the single Tauc-Lorentz oscillator entails the implementation of the following equations [45]:

$$n = \sqrt{\frac{1}{2}\left(\sqrt{\varepsilon_r^2 + \varepsilon_i^2} + \varepsilon_r\right)} \tag{1}$$

$$k = \sqrt{\frac{1}{2}\left(\sqrt{\varepsilon_r^2 + \varepsilon_i^2} - \varepsilon_r\right)} \tag{2}$$

$$E = \frac{hc}{\lambda} \tag{3}$$

$$\varepsilon_i(E) = \begin{cases} \frac{1}{E}\frac{AE_0 C(E-E_g)^2}{(E^2-E_0^2)^2 + C^2 E^2}, & \text{for } E > E_g \\ 0, & \text{for } E \leq E_g \end{cases} \tag{4}$$

$$\varepsilon_r(E) = \varepsilon_\infty + \frac{A \cdot C \cdot a_{ln}}{2 \cdot \pi \cdot \zeta^4 \cdot \alpha \cdot E_0} \cdot \ln\left[\frac{E_0^2 + E_g^2 + \alpha E_g}{E_0^2 + E_g^2 - \alpha E_g}\right] - \frac{A}{\pi \cdot \zeta^4} \frac{a_{atan}}{E_0}\left[\pi - \operatorname{atan}\left(\frac{2E_g + \alpha}{C}\right)\right.$$

$$\left. + \operatorname{atan}\left(\frac{-2E_g + \alpha}{C}\right)\right] + 2\frac{AE_0 C}{\pi \zeta^4}\left\{E_g(E^2 - \gamma^2)\left[\pi + 2\operatorname{atan}\left(\frac{\gamma^2 - E_g^2}{\alpha C}\right)\right]\right\}$$

$$- 2\frac{AE_0 C}{\pi \zeta^4}\frac{E^2 + E_g^2}{E}\ln\left(\frac{|E-E_g|}{E+E_g}\right) + 2\frac{AE_0 C}{\pi \zeta^4}E_g \ln\left[\frac{|E-E_g|(E+E_g)}{\sqrt{(E^2-E_0^2)^2 + E_g^2 C^2}}\right] \tag{5}$$

where $h$, and $c$ are Planck's constant and speed of light, and

$$a_{ln} = (E_g^2 - E_0^2)E^2 + E_g^2 C^2 - E_0^2(E_0^2 + 3E_g^2),$$

$$a_{atan} = (E^2 - E_0^2)(E_0^2 + E_g^2) + E_g^2 C^2,$$

$$\zeta^4 = (E^2 - \gamma^2)^2 + \frac{\alpha^2 C^2}{4},$$

$$\alpha = \sqrt{4E_0^2 - C^2},$$

$$\gamma = \sqrt{E_0^2 - C^2/2},$$

After combining all the above equations, $n(\lambda)$ and $k(\lambda)$ are parameterized by five fitting parameters, $A$, $C$, $E_0$, $E_g$, and $\varepsilon_\infty$. We fix $\varepsilon_\infty = 0$, and sample grids for each parameter as follows—$A$, 10 to 200 with 11 grid nodes; $C$, 0.5 to 10 with 10 grid nodes; $E_0$, 1 to 10 with 10 grid nodes, and $E_g$, 1 to 5 with 10 grid nodes. After sampling, we randomly select 1,116 $n$, $k$ spectra to be included in our dataset.

**Transfer-Matrix Method Simulation**

The Python implementation of TMM simulation follows the equations in [48], assuming no roughness, and fully coherent layers. The incident angle is 0°, and the incident medium above the films and the exit medium below the glass substrate are air (with infinite thickness). The glass substrate with a thickness of 1 mm corresponds to the actual substrate used in depositing the six MAPbI$_3$ films. The incident angle around 0° for transmission and 8° for reflectance is also used in the UV-Vis measurement due to the setup of the integrating sphere. The small discrepancy of incidence angles between measurement and simulation only causes negligible difference in reflectance spectra.

**Convolutional Neural Network**

Designed for image recognition, the classic CNN architecture consists of three main types of layers: convolutional, pooling and fully connected. Convolutional layers connect with input and each other through local filters of fixed sizes and extract the features within the filter window through convolution. In the convolutional layer, information from local pixels (within the filter window), or the local feature, gets concentrated and to be passed as a single pixel to the next layer. With the addition of each convolutional layer, features extracted are of higher and higher level (features that are more global). Thus, the series of convolutional layers becomes a feature extractor, containing features ranging from low level to high. A pooling layer usually follows convolutional layers, to downsample the spatial dimensions of the given input. Max pooling (retaining the maximum values during downsampling) has widest usage, which aims to retain the most salient features. Fully connected layers are exactly a multilayer perceptron (MLP), taking the extracted features as input. The name "fully connected" arises from the comparison with convolutional layers, which are locally connected through filters. An MLP is a universal approximator [49] for mappings between input and output, and serves to learn the mapping from extracted features to the output. In addition to the three types of layers, *thicknessML* also adds dropout layers after fully connected layers to prevent overfitting. The detailed layers and hyperparameter of *thicknessML* can be found in section S1 of Supplementary Information.

Section S2 of Supplementary Information peeks into the black box of *thicknessML*, and visualizes activation maps of an example $R$, $T$ spectra (from the source dataset). The four rows of activation maps correspond to the outputs of the four convolutional layers (after ReLU activation) respectively (ten filters are randomly chosen for each convolutional layer to produce the activation maps). Certain filters activate maximally at peaks or valleys of the $R$, $T$ spectra, which closely relates to film thickness.

**Multitask Learning**

Multitask learning is concurrent learning of multiple tasks, while each task can be a regression or classification task as in supervised learning. This concurrent learning is achieved by parameter sharing, which can be implemented via hard (using the same parameters) or soft (using similar parameters) parameter sharing. Ruder provides a helpful overview of multitask learning in [42]. *thicknessML* adopts the hard parameter sharing, letting the prediction of $d$, $n$, and $k$ share the same parameters of convolutional layers, *i.e.*, the same feature extractor. The shared feature extractor promotes extraction of more generalized features, and the auxiliary tasks also provide regularization by introducing an inductive bias. The three tasks of $d$, $n$, and $k$ prediction retain individual fully connected layer blocks to individually process and map the extracted features to respective values.

**Transfer Learning**

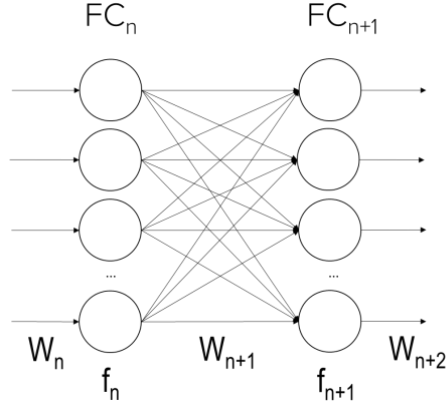

Figure 6. A simplified neuron representation of fully connected layer n and n+1, where $W_n$ denotes the weights associated with layer n, and $f_n$ the activation functions.

Figure 6 depicts a simplified representation of certain fully connected layers with associated weights and activation functions. Weights are to perform weighted sum with incoming inputs from previous layers, and activation functions are to decide whether to activate with a hard or soft cut-off based on the weighted sum. The convolutional layers follow the same principle except the incoming inputs are spatially arrange, and the weights are in the spatial form of filters. During the pre-training, and learning from scratch of *thicknessML*, the weights are randomly initialized, and the weights are updated through training data via backpropagation. The knowledge of an inverse mapping of TMM of an underlying TL oscillator is embedded in the trained weights of the pre-trained *thicknessML*. This knowledge via the pre-trained weights is then transferred to the perovskite target dataset in two manners—continue to update only the weights of the fully connected layers (partial-weight retraining), or weights of both the convolutional layers and the fully connected layers (full-weight retraining).

**MAPbI$_3$ Film Deposition, UV-Vis Measurement, and Profilometry Measurement**

In the deposition of MAPbI$_3$, six combinations of two thickness-affecting process variables, concentration of the perovskite precursor solution (PbI$_2$ and MAI with molar ratio of 1:1), and spin coating speed, are used and recorded in Table 3. The deposited films are then measured for UV-Vis with an Agilent Cary 7000 UV-Vis-NIR Spectrophotometer, and for profilometry with a KLA Tencor P-16 + Plus Stylus Profiler.

Table 3. Values of concentration of precursor solution and spin coating speed used in the deposition of MAPbI$_3$ films, and measured and predicted thickness of the films (as recorded in Figure 5).

| Film No. | Concentration of Precursor Solution (M) | Spin Coating Speed (rpm) | Measured Thickness (nm) | Predicted Thickness (nm) |
|---|---|---|---|---|
| 1 | 0.5 | 3000 | 154.17 | 122.8 |
| 2 | 0.5 | 6000 | 99.29 | 101.3 |
| 3 | 1.25 | 3000 | 389.89 | 418.5 |
| 4 | 1.25 | 6000 | 265.07 | 256.0 |
| 5 | 1.5 | 3000 | 460.35 | 489.9 |
| 6 | 1.5 | 6000 | 311.15 | 373.8 |

**Data Availability**

Datasets used in this study is provided in https://github.com/PV-Lab/thicknessML.

**Code Availability**

Code for pre-training, and transfer learning is provided, together with pre-trained *thicknessML* models, in https://github.com/PV-Lab/thicknessML.

**Acknowledgements**

We acknowledge financial support from the National Research Foundation (NRF) Singapore through the Singapore Massachusetts Institute of Technology (MIT) Alliance for Research and Technology's Low Energy Electronic Systems research program, and the Energy Innovation Research Program (grant number, NRF2015EWT-EIRP003-004 and NRF-CRP14-2014-03 and Solar CRP: S18-1176-SCRP), TOTAL SA research grant funded through MITei, the Accelerated Materials Development for Manufacturing Program at A*STAR via the AME Programmatic Fund by the Agency for Science, Technology and Research under Grant No. A1898b0043, and Solar Energy Research Institute of Singapore (SERIS), a research institute at the National University of Singapore (NUS) supported by the National University of Singapore (NUS), the National Research Foundation Singapore (NRF), the Energy Market Authority of Singapore (EMA), and the Singapore Economic Development Board (EDB). C.P.B., B.P.M., and F.G.L.P. are grateful to the Canadian Natural Science and Engineering Research Council (RGPIN-2018-06748) and Natural Resources Canada's Energy Innovation Program (EIP2-MAT-001) for financial support. QL is supported by the National Research Foundation, Singapore, under the NRF fellowship (NRFNRFF13-2021-0005).

**Author Contribution**

Conceptualization, B.P.M., F.G.L.P., F.O., C.P.B. and T.B.; Methodology, Z.L., S.I.P.T., Z.R., T.B., Q.L., Y.F.L., J.S., F.O., B.P.M. and F.G.L.P.; Software, Z.L. and S.I.P.T.; Investigation, S.V., Y.C., D.B., V.C., and S.I.P.T.; Writing – Original Draft, S.I.P.T., and Z.L.; Writing – Review & Editing, Z.L., T.B. and S.I.P.T.; Funding Acquisition, T.B. and A.G.A.; Resources, A.G.A.; Supervision, Z.L., T.B., and Q.L.

**Competing interests**

Although our laboratory has IP filed covering photovoltaic technologies and materials informatics broadly, we do not envision a direct COI with this study, the content of which is open sourced. Three of the authors (Z.R., T.B., Z.L.) own equity in Xinterra Pte Ltd, which applies machine learning to accelerate novel materials development.

# Supplementary Information

## S1. *thicknessML* Hyperparameters

Table S1. *thicknessML* hyperparameters

| *thicknessML* Hyperparameters | Setting |
|---|---|
| Convolutional layers | 4 |
| Filter size | [8, 5, 3, 3] |
| No. of filters | [512, 128, 64, 32] |
| Max pooling layers | 4 |
| Spatial extent of pooling layers | [3,3,2,2] |
| Number of units for fully connected layers for $d$ | [2048, 1024, 512, 1] |
| Number of units for fully connected layers for $n$ | [2048, 1024, 651] |
| Number of units for fully connected layers for $k$ | [2048, 1024, 651] |
| Dropout rate of fully connected layers | 0.3 |
| Epochs | 2000 |
| Batch size | 128 |
| Optimization algorithm | AdaGrad |
| Learning rate | 0.001 (initial and restored every 50 epochs since 150) |

## S2. Visualization of *thicknessML* Activation Maps of an Example $R$, $T$ Spectra

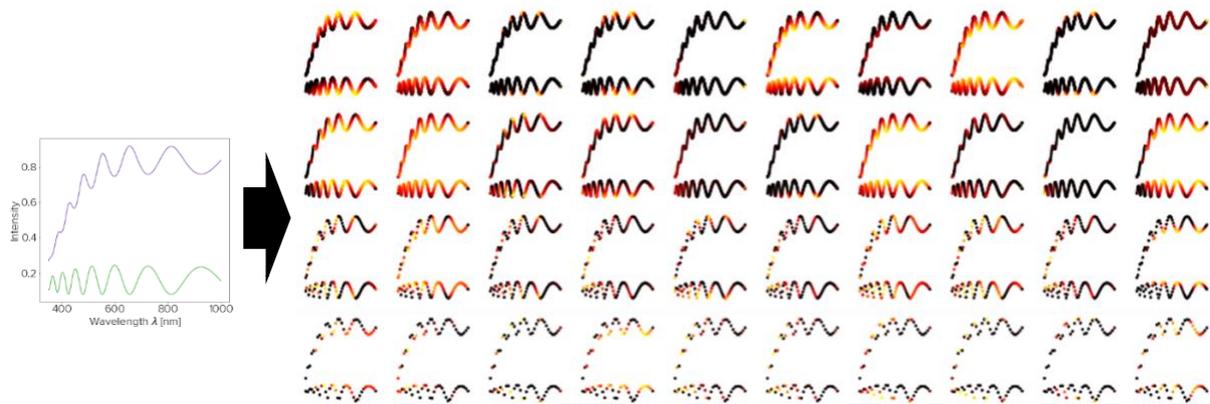

Figure S1. Visualization of activation maps of ten random filters of the 4 convolutional layers (top to bottom)
First to fourth convolutional layer) from a pre-trained *thicknessML*. Lighter color indicates higher activation of the spectra.

Figure S1 plots the activation maps of an example $R$, $T$ spectra, where an activation map is the output of a given filter (weights of convolutional layers) applied to the previous layer. Activation maps give some intuition into which part of the spectra each filter focuses on (gets more activated). For instance, we can see some filters focus on extracting the hill tops and valley bottoms of the spectra, some filters get activated when encountering certain gradients like uphill or downhills portions, and some filters recognize more holistically where the whole hills and valleys are. This tallies well with the physical picture, where positions of the hills and valleys are the most prominent feature in the $R$, $T$ spectra when thickness $d$ is varied. More oscillations (hills and valleys) are being seen with a larger $d$.